\documentclass[10pt,twocolumn,letterpaper]{article}

\usepackage[pagenumbers]{cvpr} 

\usepackage{float} 
\usepackage{dblfloatfix}
\usepackage{graphicx}
\usepackage{caption}

\definecolor{cvprblue}{rgb}{0.21,0.49,0.74}
\usepackage[pagebackref,breaklinks,colorlinks,allcolors=cvprblue]{hyperref}

\def\paperID{*****} 
\def\confName{CVPR}
\def\confYear{2025}

\title{DecoDINO: 3D Human-Scene Contact Prediction with Semantic Classification}

\author{
Lukas Bierling\\
\and
Angelo Broere\\
\and
Fleur Dolmans\\
\and
Helia Ghasemi
\and
Davide Pasero \\
\and
  University of Amsterdam \\
  \texttt{\small \{lukas.bierling, fleur.dolmans, helia.ghasemi, davide.pasero\}@student.uva.nl}
}

\begin{document}
\twocolumn[{
  \renewcommand\twocolumn[1][]{#1}

  \maketitle

  \begin{center}
    \vspace{-0.5cm}
    \includegraphics[width=0.9\textwidth,
                     height=0.60\textheight,
                     keepaspectratio]{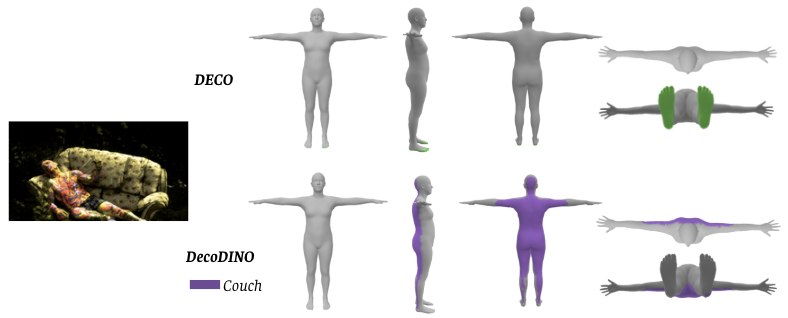}
    \captionof{figure}{DecoDINO improves DECO's performance on infering better dense vertex-level 3D contacts on the full human body. Given an RGB image, DecoDINO captures better binarry contact, handles failure cases (e.g. occlusion) and class imbalance (e.g. false foot contact  prediction) better. Additionally, it enhances DECO with semantic classification, allowing DecoDINO to predict that the contact object is a couch.}
    \label{fig:teaser}
  \end{center}
}]



\begin{abstract}
Accurate vertex-level contact prediction between humans and surrounding objects is a prerequisite for high-fidelity human–object interaction models used in robotics, AR/VR, and behavioral simulation. DECO was the first in-the-wild estimator for this task but is limited to binary contact maps and struggles with soft surfaces, occlusions, children, and false-positive foot contacts. We address these issues and introduce DecoDINO, a three-branch network based on DECO's framework. It uses two DINOv2 ViT-g/14 encoders, class-balanced loss weighting to reduce bias, and patch-level cross-attention for improved local reasoning.  Vertex features are finally passed through a lightweight MLP with a softmax to assign semantic contact labels. We also tested a vision-language model (VLM) to integrate text features, but the simpler architecture performed better and was used instead. On the DAMON benchmark, DecoDINO (i) raises the binary-contact F1 score by 7$\%$, (ii) halves the geodesic error, and (iii) augments predictions with object-level semantic labels. Ablation studies show that LoRA fine-tuning and the dual encoders are key to these improvements. DecoDINO outperformed the challenge baseline in both tasks of the DAMON Challenge. Our code is available at \url{https://github.com/DavidePasero/deco/tree/main}.
\end{abstract}    
\section{Introduction}
\label{sec:intro}

 Predicting and understanding physical contact between humans and objects in images is fundamental for modeling realistic human-object interactions (HOI) and human-scene interaction (HSI). This capability is crucial for downstream applications in robotics, virtual and augmented reality, and human behavior simulation. Knowing which object is touched as well (i.e., semantic classification) as where contact occurs further improves downstream performance \citep{cao2023detectinghumanobjectinteractionrelationship}.
 
 DECO (Dense Estimation of 3D Human-Scene Contact) \citep{tripathi2023deco} is one of the first methods to infer vertex-level binary contact on a 3D SMPL body mesh \citep{loper2023smpl} from a single RGB image. By reasoning over body pose, proximity, and scene context, DECO focusses on binary prediction: "contact" vs. "no contact". However, it provides no semantic class of the contacted object (e.g., floor vs table) which restricts downstream tasks that require detailed contextual understanding of interactions \citep{cao2023detectinghumanobjectinteractionrelationship}. Additionally, DECO fails in occlusion-rich scenes and especially producing systematic false-positive foot contacts. A detailed qualitative analysis of these errors is crucial to reveal shortcomings in both DECO’s visual features and its loss design to get a better understanding.
 
 Recent advances in self-supervised feature learning, particularly the DINOv2 vision transformer, show significant improvement in extracting powerful, general-purpose visual representations \citep{oquab2023dinov2}. It produces task-agnostic representations that are applicable and highly effective for a variety of Computer Vision tasks, including pixel-level and dense prediction tasks \citep{oquab2023dinov2}. These properties suggest that DINOv2 features could benefit both binary contact prediction and semantic classification.
 
In this study, we contribute to three key objectives: (1) conducting a qualitative analysis of DECO to better understand its failure modes, (2) improving the performance and robustness of binary contact prediction, and (3) introducing semantic classification of contacted objects. To this end, we present DecoDINO, which retains DECO’s overall structure but replaces its encoder with two pretrained DINOv2-Giant models, one focusing on global scene context and the other on local body-part context. Both encoders are adapted with Low-Rank Adaptation (LoRA) for parameter-efficient fine-tuning \citep{hu2022lora}. Further, to address common failure cases (e.g. such as occlusions and persistent false-positive foot contact predictions) we introduce a positive class balance weight to the loss function that mitigates the effects of class imbalance in the training data. Additionally, we replace the class-level cross-attention mechanism with a patch-level attention module to capture more fine-grained contextual information.

This work is carried out within the scope of the RHOBIN Challenge \footnote{\url{https://rhobin-challenge.github.io/index.html}}, a CVPR 2025 workshop co-organized by UvA. We contribute to two challenges evaluated on the DAMON test set; binary contact prediction\footnote{\url{https://codalab.lisn.upsaclay.fr/competitions/21775\#results}} and semantic contact classification\footnote{\url{https://codalab.lisn.upsaclay.fr/competitions/21781}}.



\subsection{Related work}
DECO \citep{tripathi2023deco} was one of the first methods to infer dense, per-vertex binary contact between a human and surrounding scene objects from a single RGB image and project it onto a SMPL body mesh \citep{loper2023smpl}.

 LEMON (LEarning 3D huMan-Object iNteraction relation) \citep{yang2024lemonlearning3dhumanobject} is a unified model that jointly predicts multiple interaction elements by minimizing geometric correlations via surface curvatures and learning interaction intentions from 2D images. While DECO focuses solely on binary contact labels per SMPL vertex, LEMON expands the scope by also predicting object-centric affordance regions and spatial relationships, capturing a more comprehensive representation of human-scene interactions. LEMON's joint reasoning over contact and affordances highlights the potential of extending DECO with semantic classification, which could enhance both binary contact prediction and per-vertex object labeling.

\citet{cseke2025picoreconstructing3dpeople} introduce PICO-db, a dataset that extends DAMON’s 3D body-contact annotations with 3D object-contact labels. Object meshes are retrieved with vision foundation models, and body-contact patches are mapped to the objects through a two-click procedure, keeping manual input minimal. The authors also present PICO-fit, which jointly optimizes body and object geometry to the input images, enabling object-aware reconstructions across categories that earlier methods could not handle. Our study remained limited to DAMON due to the RHOBIN challenge, but PICO-db and PICO-fit are promising additions for future work.

Recently, \citet{dwivedi2025interactvlm3dinteractionreasoning} propose InteractVLM, a “Render–Localise–Lift” pipeline in which a vision-language model predicts 2D human- and object-contact points that are subsequently lifted to 3D. The use of a VLM mitigates occlusion effects and reduces annotation requirements compared with DECO. This approach motivated the integration of a VLM component in our own architecture.
 
\section{Background}
\subsection{DECO}
DECO \citep{tripathi2023deco} is designed to predict dense, per-vertex 3D human-scene and human-object contact from a single RGB image. It leverages the SMPL body mesh \citep{loper2023smpl}, containing 6890 vertices, and integrates three interacting branches: scene-context, part-context and contact branch.

In the scene-context and part-context branches, a scene encoder $E_{s}$ and a part encoder $E_p$ extract scene features $F_s$ and body-part features $F_p$, respectively. These encoders are trained to identify relevant visual features by utilizing a corresponding scene decoder $D_s$ and part decoder $D_p$. Specifically, $D_s$ outputs semantic segmentation maps, over MS-COCO object categories \citep{lin2014microsoft}, while $D_p$ produces a 25-channel part segmentation (24 SMPL body parts + background class). 

Within the contact branch, extracted scene and part features are fused through a cross-attention mechanism. This approach enables each branch to attend to relevant regions from the other branch's features ($F_s$ and $F_p$). The cross-attention results are combined using element-wise multiplication (Hadamard product) and layer normalization, and subsequently processed by a multi-layer perceptron (MLP) with sigmoid activation to produce vertex-level contact probabilities $\bar{y}_c$ on the SMPL mesh.

DECO is trained end-to-end using a composite loss function $\mathcal{L}$ (Eq. \ref{total_deco_loss}). This loss consists of binary cross-entropy loss $\mathcal{L}_c^{3D}$ (Eq.\ref{BCE}) between predicted vertex-level contacts and ground-truth contacts, scene and part segmentation losses comparing predicted and ground-truth segmentation masks, and a pixel anchoring loss that aligns 3D mesh predictions with image pixels. A detailed architecture description, including further explanations of the cross-attention mechanism and loss function, is provided in Appendix \ref{DECOappendix}.

\subsection{DINOv2}
DINOv2 \citep{oquab2023dinov2} builds upon the standard Vision Transformer (ViT) backbone (with a patch size of 14), enhancing it with two parallel self-supervised objectives operating at different granularities: image-level (DINO) and patch-level (iBOT). This self-supervised method combines aspects of DINO \citep{caron2021dino} and iBOT \citep{zhou2021ibot} losses, further refined using a centering strategy inspired by SwAV \citep{caron2020SwAV}. 

At the image-level, DINO employs a student-teacher framework where class tokens from differently cropped image views feed into separate multi-layer perceptron (MLP) heads, generating a vector of "prototype scores". These scores undergo softmax normalization to form student logits $p_s$ and teacher logits $p_t$, with the teacher logits additionally centered using either moving averages or Sinkhorn-Knopp normalization. 

At the patch level, iBOT involves masking some input patches presented to the student model, while the teacher model receives the unmasked patches. Both student and teacher heads produce logits for corresponding patches, with the teacher logits centered similarly as in the DINO approach. See Appendix \ref{DINO} for a formal notation of the losses.

\section{Methodology}
We begin by qualitatively analyzing the failure modes of DECO to better understand its limitations. The insights gained from this analysis inform the design of DecoDINO.

\subsection{Qualitative analysis of DECO}
\label{sec: 3.1}
In the analysis we investigate DECO's performance in four stages: (1) investigate class imbalance in DECO's training datasets,
(2) evaluated performance under challenging scenarios, (3) visual inspection of scene and part segmentations, and (4) ablation by zeroing out features. 

For this, we compiled 16 images from the DAMON test set and 4 from Google featuring different challenging scenarios: no foot contact, soft materials, occlusions, cropped bodies, and children. The images are passed as input to DECO, which predicts the binary contact on SMPL body meshes. Additionally, we visualize the predicted part and scene mask from the part and scene branches to get a better understanding of where the model fails. We provide an interactive notebook to see the all qualitative analysis\footnote{\url{https://github.com/DavidePasero/deco/Qualitative}}. In Appendix \ref{Qualitative}, Figs. \ref{fig:part1}-\ref{fig:part2} are some of these predictions and masks shown. 

\paragraph{Class imbalance.}
DECO was trained on the DAMON \citep{tripathi2023deco}, RICH \citep{huang2022capturing}, and PROX \citep{hassan2019resolving} datasets. To assess the distribution of contact in these datasets, we inspected all training and validation images and counted frames with at least one contact vertex. In DAMON and RICH, only 0.2\% of the images lack contact entirely, while in PROX, 6.1\% of the images contain less than one contact vertices. This indicates that the model is rarely exposed to contact-free examples in DAMON and RICH, but encounters them more frequently in PROX.

\begin{figure}[!h]
    \centering
    \includegraphics[width=1\linewidth]{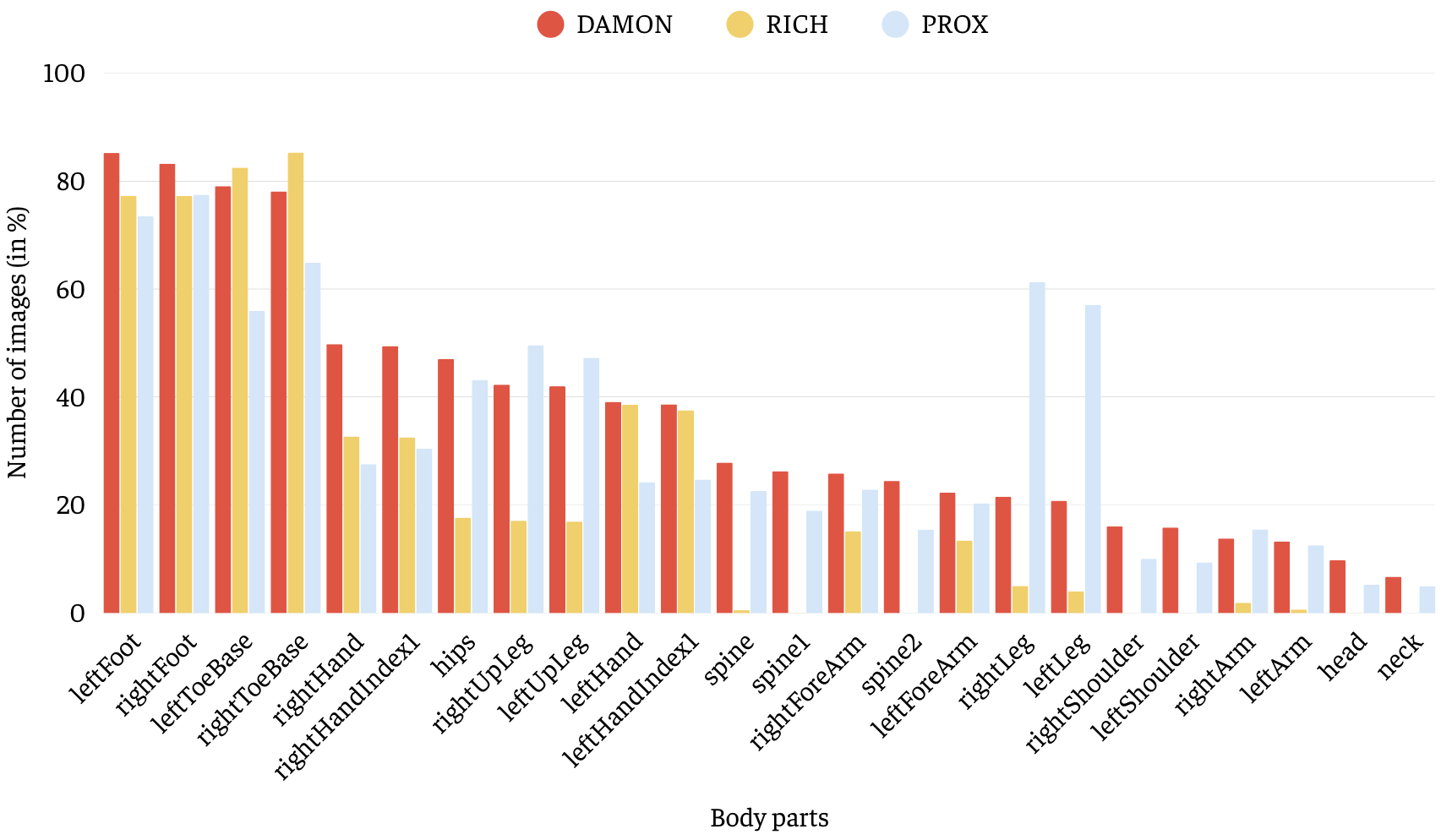}
    \caption{Number of images in the DAMON dataset with contact per body part. A body part is counted if any of its vertices are in contact.}
    \label{fig:imbalance}
\end{figure}

Figure~\ref{fig:imbalance} shows the distribution of contact across body parts in DAMON. A body part is considered in contact if at least one of its vertices is labeled as such. A cross all three datasets, contact is dominated by the feet with more than 80\% of the images having any feet contact. In RICH, the left and right hands also occur frequently, while in DAMON, the right hand is commonly in contact. In PROX, both legs appear prominently. In contrast, the head, arms, shoulders, and especially the neck have relatively few contact instances across all datasets. Notably, spine and shoulder contact labels are absent in RICH.

This imbalance is expected given the effect of gravity; i.e. individuals are typically in contact with the ground through their feet during common activities such as standing, sitting, or walking. Such imbalance is likely not limited to DAMON, RICH, and PROX but are likely also present in HSI and HOI datasets. As a result, models trained on these datasets may develop a bias toward predicting contact for frequently occurring regions, such as the feet and hands, while failing to generalize to less commonly involved body parts.

\begin{figure}[!h]
    \centering
    \includegraphics[width=1\linewidth]{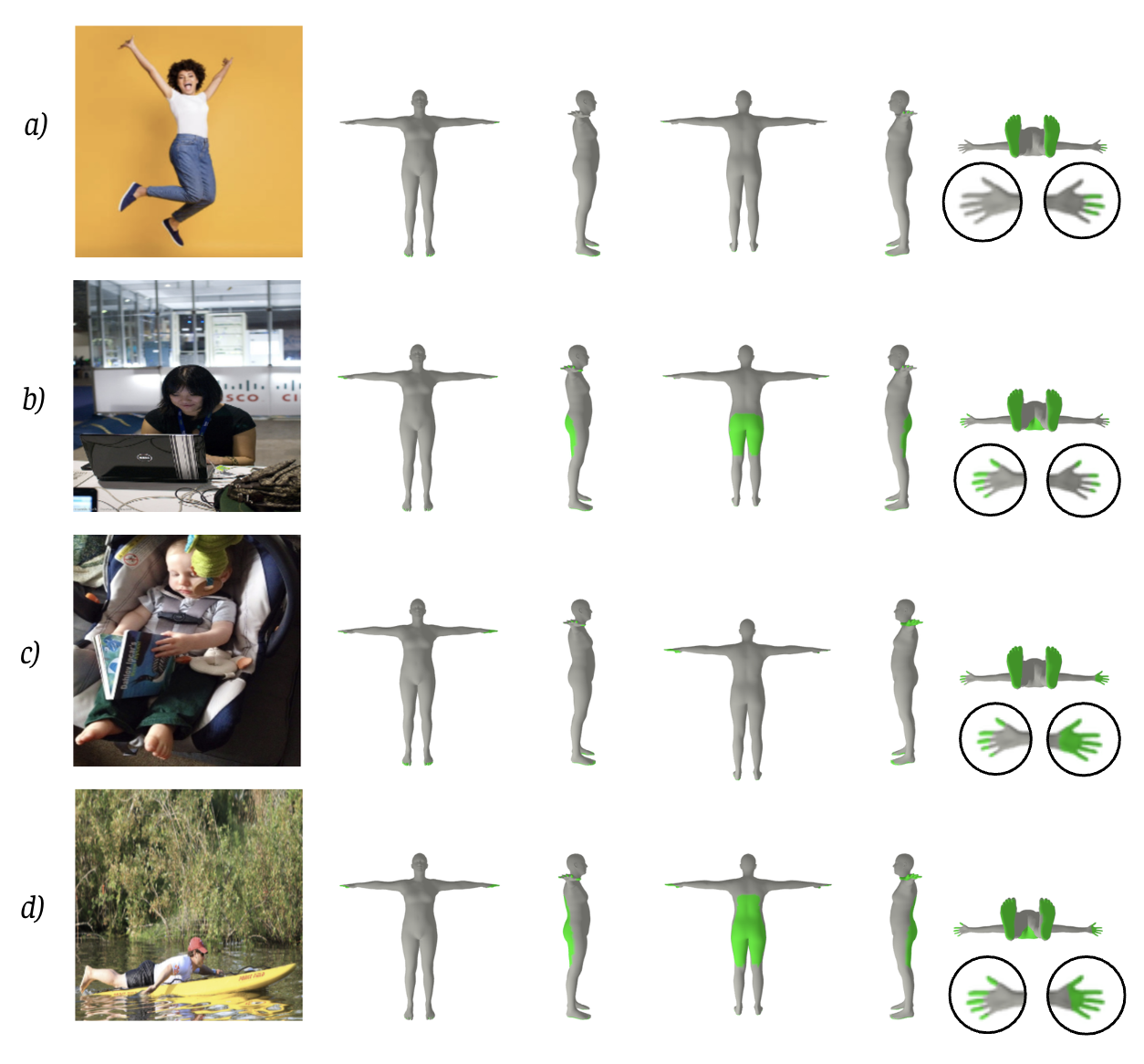}
    \caption{DECO's binary contact prediction on challenging scenario's}
    \label{fig:Challenging scenario's}
\end{figure}

\paragraph{Challenging scenarios.} Fig. \ref{fig:Challenging scenario's} illustrates three examples of DECO failure modes. The most common error is systematic false positive foot contact, where the model predicts ground contact even when the subject’s feet are clearly off the ground, such as during jumps or while lying down (see panels a, c and d). This tendency is largely attributable to the class imbalance. As a result, the model develops a bias toward predicting foot contact in diverse scenarios. In contrast, performance in cropped images of seated or standing adults is satisfactory (panel b), where contacts on thighs and buttocks are more reliably detected. Performance drops for children; the model occasionally identifies hand- or foot-ground contacts but almost never flags contact by other body parts (panel c). Also, lying poses are particularly challenging; the model misses to predict any contact (except for false feet contact) or, for example, predicts contact on the wrong side of the body (panel d). Soft surfaces (e.g., couches) and occlusions further degrade predictions, with contact regions consistently underestimated.

These findings suggest that DECO struggles to generalize beyond its training distribution, particularly in cases with atypical poses, occlusions, or subjects such as children. The systematic over-prediction of foot contact highlights the need for more balanced training data and improved handling of rare scenarios. Enhancing the model's robustness may require targeted data augmentation, explicit handling of soft materials, increased diversity in annotated poses and subjects and introducing weights.

\paragraph{Scene and part segmentations.}
 The qualitative results show that both branches generally produce weak segmentations. Performance tends to degrade in seated or recumbent poses; for example, the model sometimes mislabels an entire sofa as a person or fails to detect the body entirely, resulting in an empty part mask. Scene masks are more reliable when the subject is standing and unobstructed, though even than, the segmentation boundaries remain coarse. Hands and feet are frequently missing from the part masks.

 \begin{figure}[!h]
     \centering
     \includegraphics[width=1\linewidth]{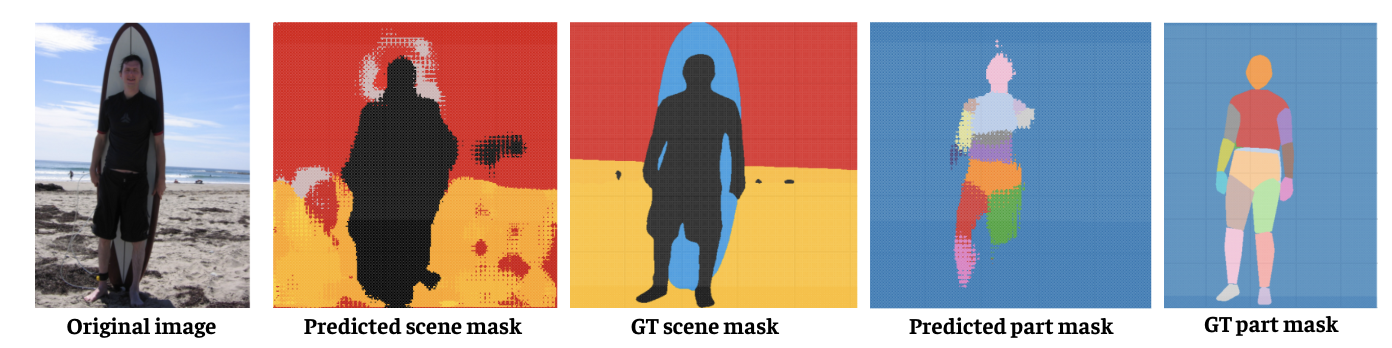}
     \caption{Predicted scene and part segmentation with their ground truth (GT). \textbf{Note}: Predicted and ground-truth part masks use different label colors but refer to the same body parts (e.g., the head appears orange in the ground truth and pink in the prediction).}
     \label{fig:Scene and part}
 \end{figure}

To assess whether DECO has learned meaningful part representations, we compared its part predictions on several DAMON images against their ground-truth masks. One example is shown in Fig. \ref{fig:Scene and part}, in which hands and feet are again absent from the predicted part mask. Additionally, the scene mask includes several misclassified or imprecise regions. Despite these visual differences, the predicted parts generally correspond to the correct classes in the ground truth, indicating that the class embeddings are meaningful. Still, the frequent omission of body parts and the variable quality of the scene mask suggest that the scene and part context modules contribute only limited discriminative value to the final predictions.

\begin{figure}[!h]
     \centering
     \includegraphics[width=1\linewidth]{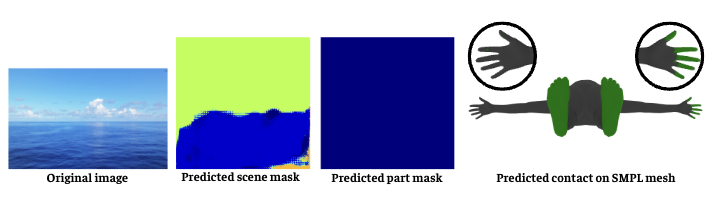}
     \caption{\textbf{Scene without a person.} Predicted scene and part segmentation with the contact prediction on a SMPL body mesh on a scene without a person.}
     \label{fig:no person}
 \end{figure}

\paragraph{Zeroing-out features.}
To assess the model’s reliability, we evaluated its performance on images that do not contain a person (example shown in Fig. \ref{fig:no person}).

\begin{figure}[!h]
    \centering
    \begin{subfigure}[t]{0.1\linewidth}
        \centering
        \includegraphics[width=1cm]{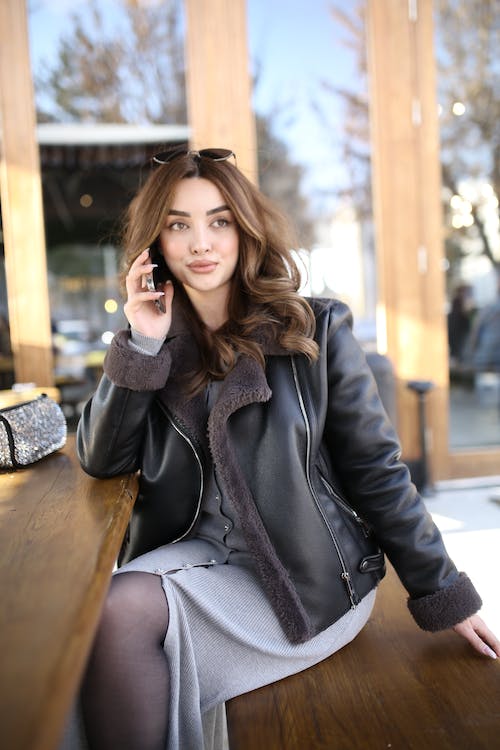}
        \caption{}
        \label{input}
    \end{subfigure}%
    \begin{subfigure}[t]{0.2\linewidth}
        \centering
        \includegraphics[width=1.5cm]{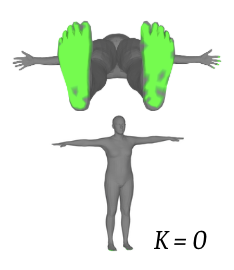}
        \caption{}
        \label{k0}
    \end{subfigure}%
    ~ 
    \begin{subfigure}[t]{0.2\linewidth}
        \centering
        \includegraphics[width=1.6cm]{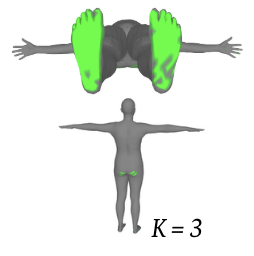}
        \caption{}
    \end{subfigure}
    \begin{subfigure}[t]{0.2\linewidth}
        \centering
        \includegraphics[width=1.5cm]{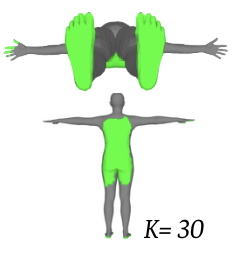}
        \caption{}
    \end{subfigure}
    \begin{subfigure}[t]{0.2\linewidth}
        \centering
        \includegraphics[width=1.5cm]{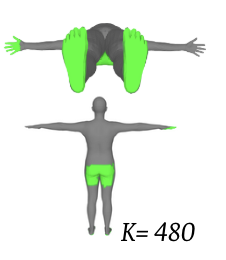}
        \caption{}
    \end{subfigure}
    \caption{\textbf{Contact prediction with zeroing-out features of the scene branch.} (a) represents the input image and (b-e) the contact predictions for different $K$, which is the number of non zero channels in the feature maps.}
     \label{fig:zero}
\end{figure}

This reveals that the model correctly outputs an empty part mask, indicating that no person is present in the image. However, it still predicts contact at the feet and a small part of the left hand on the SMPL body mesh. This suggests that DECO’s contact predictions are strongly influenced by learned priors, rather than purely by visual evidence. To further investigate this effect, we removed all scene-context information from the model. Specifically, in the scene-branch feature map $F_s \in \mathbb{R}^{H\times W\times C}$, we set $C-K$ channels to zero before the cross-attention module, leaving $K$ non-zero channels. The body-part features $F_p$ remained unchanged. Fig. \ref{fig:zero} illustrates DECO's binary contact predictions for various values of $k$. Notably, as shown in Fig. \ref{k0}, the network continues to predict foot contact even when no scene cues are present. This proves that DECO has internalized a strong "feet-on-ground" prior, a direct consequence of class imbalance in the training set.

\begin{figure*}[!h]
    \centering
    \includegraphics[width=\textwidth,
                     height=0.9\textheight,
                     keepaspectratio]{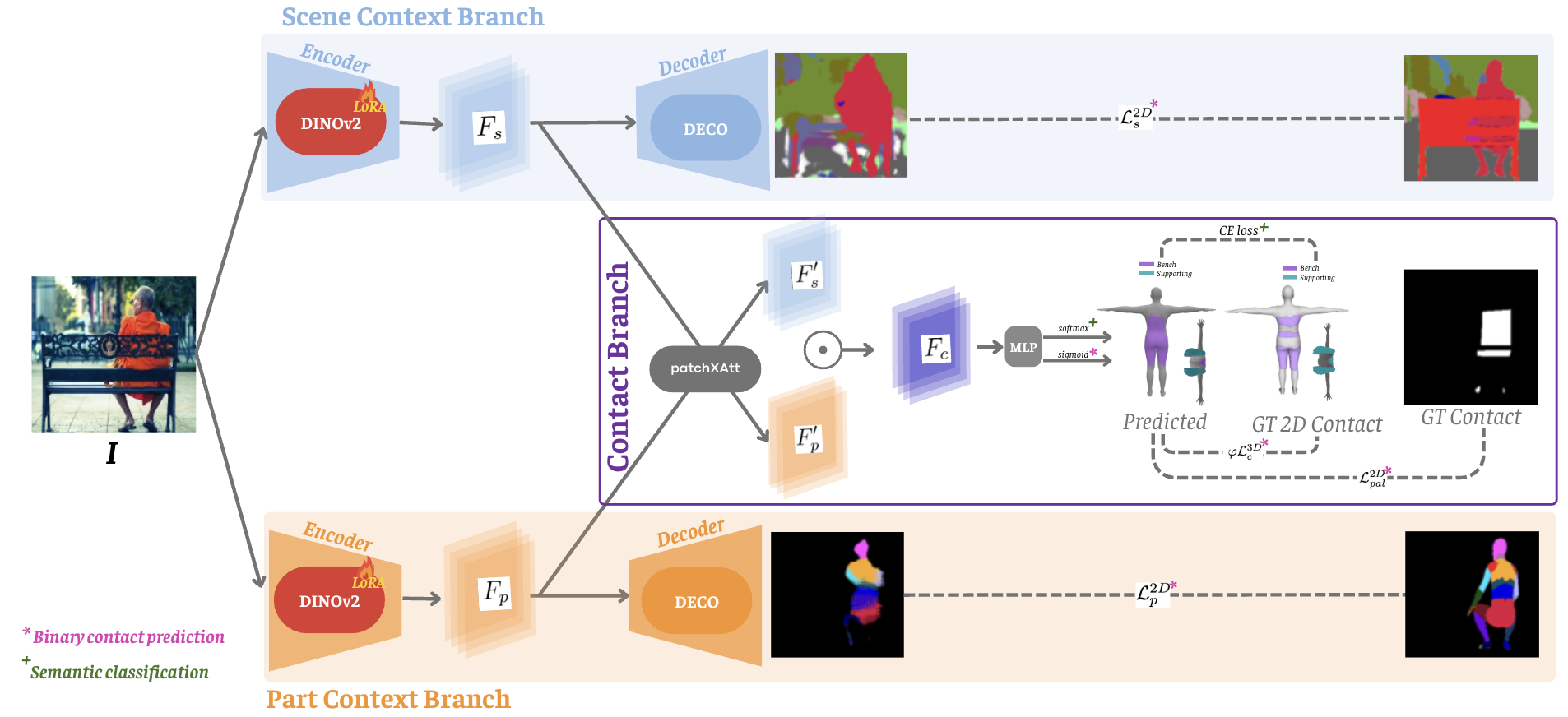}
    \caption{\textbf{DecoDINO architecture}. A single RGB image is processed by two LoRA-tuned DINOv2-G/14 encoders: a scene branch that yields scene features $F_s$ and a body-part branch that produces part features $F_p$. Both feature maps are fed to DECO's decoder to obtain scene and part segmentation masks. In parallel, a patch-level cross-attention module (patchXAtt) fuses $F_s$ and $F_p$ into a joint representation $F_c$ utilizing learned attention pooling. Finally, a MLP maps $F_c$ to vertex-level contact probabilities on the SMPL body mesh and semantic object labels, enabling both binary contact prediction and semantic classification.}
    \label{fig:dinodeco}
\end{figure*}

\subsection{DecoDINO}

\paragraph{Model Architecture.}
Fig. \ref{fig:dinodeco} depicts the DecoDINO architecture. Similar to DECO, we use three branches: scene-context, part-context  and contact branch. Given an image $\mathbf{I} \in \mathbb{R}^{H\times W \times3}$, in the scene-context and part-context branches, a scene and part encoder extract scene features $F_s$ and body-part features $F_p$, respectively. For these two encoders, we use two separate DINOv2-Giant vision encoders (ViT/g-14) which are finetuned with LoRA \citep{hu2022lora}. The features $F_s$ and $F_p$ are past to the corresponding scene and part decoder from DECO. Similarly to the original DECO framework, we pass $F_s$ and $F_p$ to a contact branch. Patch-level cross attention enables us to achieve more localized and detailed reasoning between $F_s$ and $F_p$, which is essential for accurately modeling fine-grained contact patterns. The outputted features $F_c$ are processed by multi-layer perceptron (MLP) which produces with sigmoid activation vertex-level binary contact probabilities on the SMPL mesh and classifies with a simple softmax semantic object labels. This allows us to enrich semantic classification with binary contact prediction. 

DecoDINO's binary contact prediction is trained end-to-end using DECO's original four-component loss function $\mathcal{L}$, with the addition of a positive class balance weight $\varphi$ to mitigate the over-prediction of feet contact caused by class imbalance. 

\paragraph{Positive Class Balance Weight.} 

To address class imbalance, we introduce a per-vertex positive class balance weight $\varphi$, which increases the loss contribution of rarely contacted vertices to balance out over predicted vertices. The positive weight for vertex $i$ is defined as
\begin{equation}
    \varphi_i = \frac{1}{(\frac{1- \beta^{n_i}}{1- \beta}) + \epsilon}
\end{equation}

where $n_i$ is the number of times that vertex $i$ is labeled positive, $\beta \in (0,1)$ (we use 0.99), and $\epsilon = 10^{-8}$ ensures numerical stability. To normalize the scale, the weights are rescaled to have a mean of 6.451, matching the average negative-to-positive vertex ratio. Outlier weights are clipped to prevent instability during training. This weight is added to the positive term in the binary cross-entropy loss: $ \varphi\mathcal{L}_c^{3D}$, a component of the overall DECO loss (Eq. \ref{BCE}).

\paragraph{Patch Cross-Attention.} 
In the original DECO architecture, cross-attention is computed between two vectors: the per-vertex queries of the human mesh and a global image feature, either the class token from a ViT or pooled feature maps from a convolutional encoder. While this approach is computationally efficient, it amounts to cross-attention at a scalar or global level, which discards most of the rich spatial and contextual information present in the full feature maps. True cross-attention is typically defined over sets of vectors, allowing for nuanced interactions between spatial locations in each input \cite{lin2021catcrossattentionvision}. To address this limitation, we replace the single-vector representation with patch cross-attention, which we refer to as patchXAtt. Instead of relying on the class token, we use the full set of patch embeddings from both the scene and part branches. This allows us to compute cross-attention between all patch tokens from both branches, enabling the model to reason about detailed spatial correspondences and fine-grained interactions across the entire input image. This modification enhances the model’s ability to localize and capture subtle contact patterns that might be lost when relying solely on global pooling or class-token attention. Since the binary contact classification head requires a single feature vector as input, but the patch cross-attention module produces a set of patch-level embeddings, we introduce a learned attention-based pooling mechanism. This pooling operation computes a weighted sum over the patch embeddings $F_c$, where the attention weights are learned during training. This enables the model to dynamically attend to the most informative spatial regions, improving its ability to detect fine-grained contact patterns. Further implementation details are provided in Appendix \ref{attpooling}.

\paragraph{Semantic Classification.}
To incorporate object-level context into the contact prediction framework, we extend the model with a semantic classification component. Each SMPL vertex feature in $F_c$ is passed through an MLP that performs both binary contact prediction and semantic classification. For the latter, the output is passed through a softmax layer to produce a probability distribution over predefined semantic categories, enabling per-vertex semantic labeling. These predictions are supervised using a cross-entropy loss with respect to the ground-truth semantic labels. This design results in our full DecoDINO model, which learns meaningful semantic concepts at the mesh level while leveraging improved geometric and contextual representations.
\begin{figure*}[!h]
    \centering
    \includegraphics[width=0.9\linewidth]{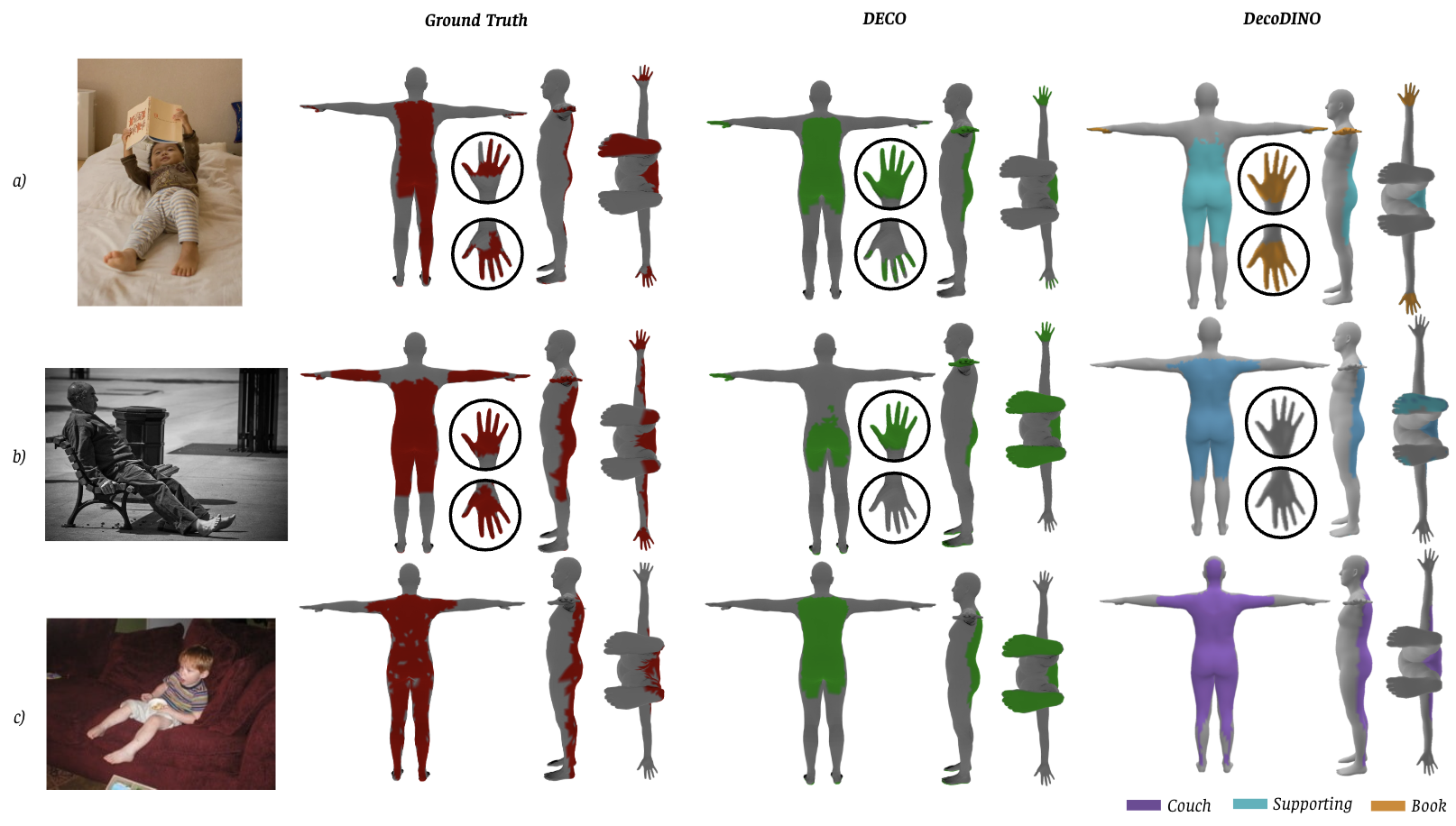}
    \caption{\textbf{Qualitative Results.} The ground truth, DECO and DecoDINO's contact prediction on SMPL body mesh. Semantic classification results for DecoDINO are shown in different colors with the corresponding legend.}
    \label{fig:results}
\end{figure*}

\section{Experiments}

\paragraph{Dataset.}

We train and test on the DAMON \citep{tripathi2023deco}, is a collection of vertex-level 3D contact labels on SMPL meshes paired with color images of people, sourced from HOT \citep{chen2023detecting}. DAMON images consist of unconstrained environments and come with both annotated human-supported contact for each individual object and scene-supported contact, retrieved from Amazon Mechanical Turk. 


\paragraph{Training and Evaluation.} We evaluate performance on the DAMON dataset using precision, recall, F1 score, and geodesic error (in centimeters) for binary contact prediction, and precision, recall, and F1 score for semantic contact classification.

\subsection{Results}
Binary contact prediction and semantic classification performances of DecoDINO are visualized in Figs. \ref{fig:results}-\ref{fig:extra} together with the performances of DECO and their ground truth. See Appendix \ref{app:qualitative results}, Fig. \ref{fig:results_extra} shows some more visualizations.

\paragraph{Binary Contact Prediction.} Tab. \ref{tab:binary} presents the performance of DECO with various incremental components added. The full combination of these components constitutes the complete DecoDINO model. 

  \begin{table}[!h]
    \centering
    \resizebox{\linewidth}{!}{%
      \begin{tabular}{lccccc}
        \toprule
      & F1$_{\%}$ & Precision$_{\%}$ & Recall$_{\%}$ & Geo. error$_{cm}$ \\
        \midrule
        DECO & 56.42  &  54.27   &   72.94  &  18.68 \\
         \quad + 2 ViT-g/14 & \textbf{63.91} $\textcolor{green}{\uparrow}$  &  58.44 $\textcolor{green}{\uparrow}$   &   \textbf{81.63} $\textcolor{green}{\uparrow}$ &  22.17 $\textcolor{red}{\uparrow}$\\
        \quad + $\varphi$  & 62.57 $\textcolor{red}{\downarrow}$ & 66.36 $\textcolor{green}{\uparrow}$ & 68.41  $\textcolor{red}{\downarrow}$ & 17.11 $\textcolor{green}{\downarrow}$\\
        \quad + patchXAtt    & 62.54 $\textcolor{green}{\uparrow}$ & \textbf{67.04} $\textcolor{green}{\uparrow}$ & 67.35 $\textcolor{red}{\downarrow}$ & \textbf{15.89} $\textcolor{green}{\downarrow}$ \\
        \bottomrule
        
      \end{tabular}
      }
      
    \caption{\textbf{Binary Contact Prediction.}. Performance of DECO and  sequently adding two LoRA-tuned ViT-g/14 encoders, a positive class imbalance weight $\varphi$ and adjusting class level cross attention to patch-level.}
    \label{tab:binary}
    
  \end{table}

Firstly, two LoRA-tuned ViT-g/14 encoders were added, already outperforms the original DECO baseline model on almost all metrices, even achieving the overall highest F1. The geodesic error or the baseline shows better performance but is very volatile with spikes to $40$ cm geodesic error in certain epochs. Subsequently, adding the positive class balance weight $\varphi$ improves precision ($+7.92\%$) and reduces geodesic error ($-5.06$ cm), indicating better localization. However, F1 and recall slightly decrease due to the nature of $\varphi$ that fewer false positives are accepted. Lastly, introducing patchXAtt slightly improves precision ($+0.68\%$) with the tradeoff that recall is a bit reduced. Geodesic error is reduced by $-1.25$ cm, suggesting more accurate spatial localization.

 Fig. \ref{fig:results}(a) shows that both models predict hand contact and absence of contact on the feet, although the ground truth indicates contact primarily with the fingers and only the right foot. DecoDINO predicts more accurate contact on the upper leg compared to DECO, but both models fail to distinguish that only one leg is in contact with the surface. In Fig. \ref{fig:results}(b), DECO significantly under-predicts contact on the back and arms but overestimates contact regions on the feet. DecoDINO captures better interactions around the torso and legs but over-predicts the shoulders and misses correct arm, hand and feet contact. In Fig. \ref{fig:results}(c), DECO misses arm and leg contact and predicts false foot contact. Unlike DECO, DecoDINO correctly predict no feet contact and more accurately captures the distributed contact areas, but overestimates head contact. 

In addition, we evaluated DecoDINO on "in-the-wild" images from Google, retrieved during the qualitative analysis, featuring scene's without contact. Fig. \ref{fig:extra} reveals that similar to DECO, DecoDINO incorrectly predicts contact on the feet. However, DecoDINO improves over DECO by not falsely predicting contact on the hands.

\begin{figure}
    \centering
    \includegraphics[width=1\linewidth]{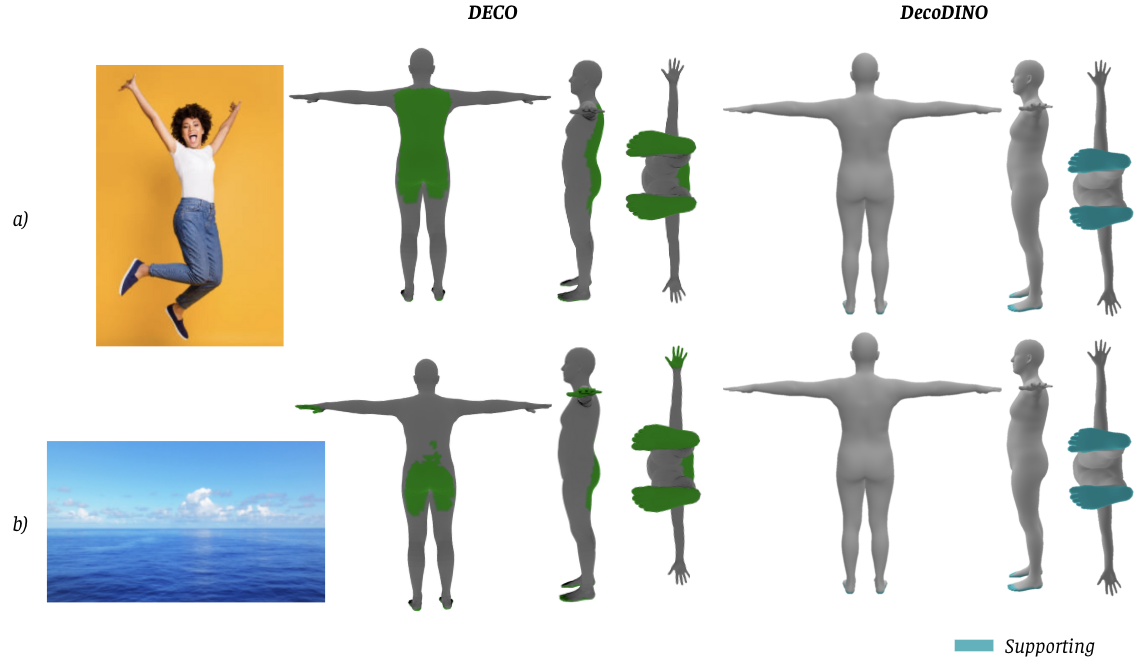}
    \caption{Performance on scene without contact.}
    \label{fig:extra}
\end{figure}

\paragraph{Semantic Classification.}To evaluate the semantic classification performance of DecoDINO, altered the original DECO with the same softmax layer to see overall model performance of both model for semantic classification. 

\begin{table*}[!b]
\centering
\resizebox{0.8\linewidth}{!}{%
    \begin{tabular}{c|cccc|ccc}
    \toprule
    &\multicolumn{4}{|c|}{\textbf{Binary Contact Prediction}} & \multicolumn{3}{c}{\textbf{Semantic Classification}}\\
      & F1$_{\%}$ & Precision$_{\%}$ & Recall$_{\%}$ & Geo. error$_{cm}$ & F1$_{\%}$ & Precision$_{\%}$ & Recall$_{\%}$\\
     \midrule
     DecoDINO (ours) &  \textbf{62.54}   &  \textbf{67.04}   &   \textbf{67.35}  &  \textbf{15.89} & \textbf{28.77} & \textbf{17.55} & \textbf{79.81}\\
    \quad + SmolVLM & 59.00   &  62.22   &   62.48  &  16.99 & 23.84  & 14.31  & 71.42 \\
    \bottomrule
    \end{tabular}
    
    }
    \caption{Effect of VLM on binary contact prediction and semantic classification performance.}
    \label{tab:vlm}
\end{table*}

\begin{table}[!h]
  \centering
    \begin{tabular}{lccccc}
    \toprule
      & F1$_{\%}$ & Precision$_{\%}$ & Recall$_{\%}$\\
      \midrule
      DecoDINO (ours)   & 28.77 & 17.55 & 79.81\\
      \bottomrule
    \end{tabular}%

  \caption{Semantic classification performance.}
  \label{tab:semantic}
\end{table}

DecoDINO’s semantic head achieves 79.8\% recall, meaning it identifies most of the relevant semantic labels. However, its precision is low at 17.55\%, indicating many incorrect positive predictions. The resulting F1 score is 28.7\%, showing an imbalanced performance with high recall but poor precision.

Fig. \ref{fig:results} demonstrates that the predicted object labels are generally reasonable. However, as illustrated in Fig. \ref{fig:extra}, the model incorrectly predicts contact in the absence of any human, labeling it as "supporting". This suggests an inductive bias in the model toward assuming that a person is present and in contact with the ground.

\subsection{VLM}
Inspired by InteractVLM, we hypothesis that incorporating a Vision Language Model (VLM) can enrich the model and improve performance. We attempted to use a VLM that could mitigate the occlusion effects and reduce annotation requirements compared with DECO. Therefore, we selected a lightweight, instruction-tuned ViT, SmolVLM-Instruct \citep{marafioti2025smolvlm}, to generate textual embeddings of human-object contacts. For each input image, we prepend the following fixed "contact-description" prompt:
\begin{quote}
    \textit{"Describe exactly which objects the human is in contact with, what action is being performed, and with what body part"}
\end{quote}
By feeding this same prompt for every frame, we ensure the VLM focuses on contact-relevant details (object categories and body parts). In  Appendix \ref{VLM prompts}, Fig. \ref{fig:prompts} are some image prompt pairs visualized. 

\begin{figure}[!h]
    \centering
    \includegraphics[width=1\linewidth]{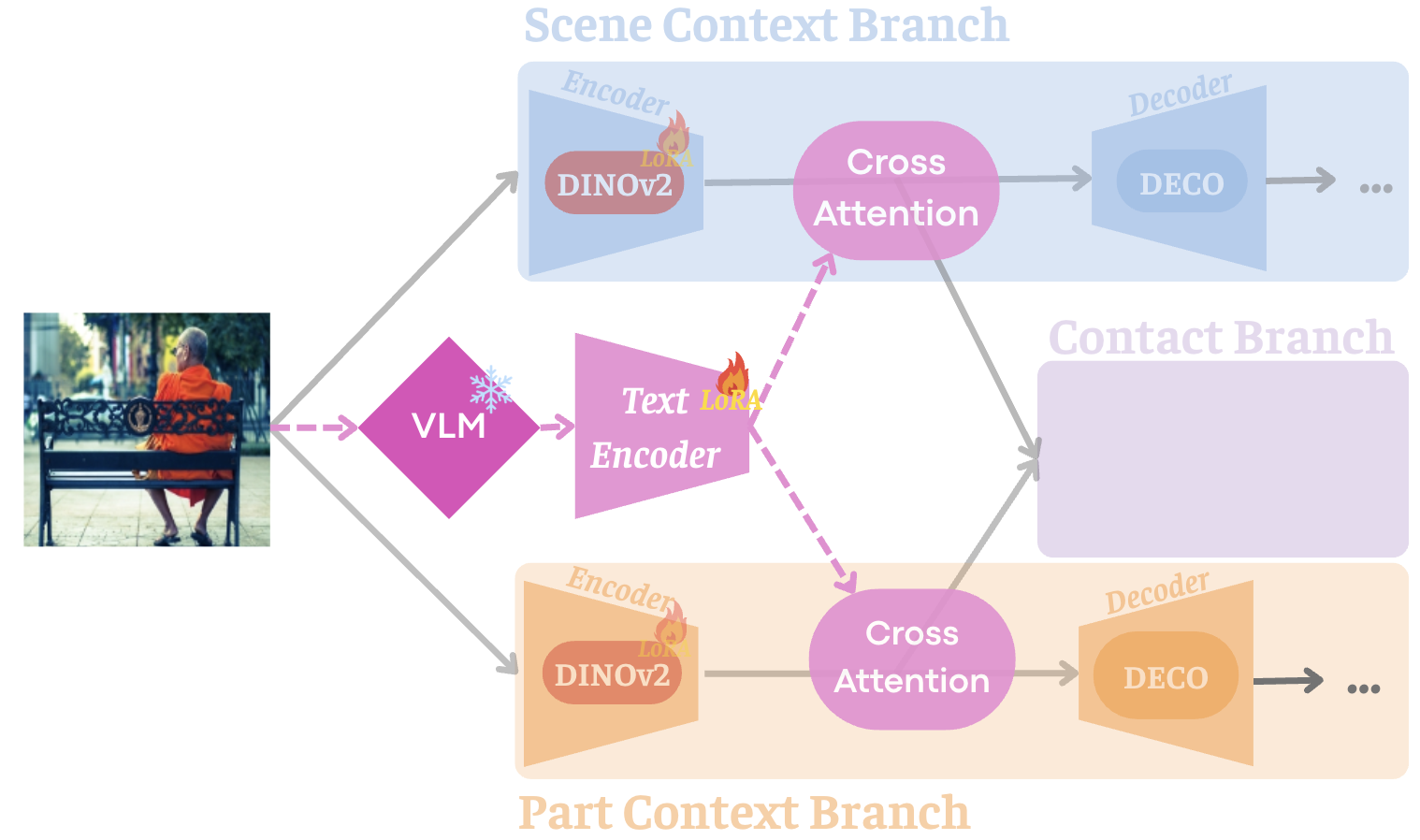}
    \caption{DecoDINO architecture adjustment when incorporating VLM}
    \label{fig:with_VLM}
\end{figure}

Fig. \ref{fig:with_VLM} shows where the VLM is incorporated into DecoDINO's architecture. It takes an image as input and generates a vector with text tokens, representing the image. These are passed to a text encoder, after which the features are inputted to separate cross-attention in the scene and part context branch to enhance the image features with text features. From there, the model will continue as the DecoDINO architecture visualized in Fig. \ref{fig:dinodeco}. Detailed inference steps are listed in Appendix \ref{VLM inference}.

\paragraph{Results.}
 
Despite our hypothesis, Tab. \ref{tab:vlm} shows that integrating a VLM does not improve performance. Compared to our DecoDINO model, the incorporation of a VLM under-performs across all metrics on the binary contact prediction and semantic classification.
This outcome suggests that, in this context, the VLM features may introduce noise or irrelevant information rather than providing meaningful context for the binary contact prediction task. It is possible that the text features are not sufficiently aligned with the fine-grained spatial cues required for accurate contact detection, or that the additional modality complicates the learning process without offering complementary information. Further analysis is needed to better understand the interaction between text-derived features and dense spatial predictions in this setting.

\vspace{0.2cm}

\subsection{Ablation Studies}

We conduct ablation studies (see Appendix \ref{ablation}) to validate key design choices. Specifically, we assess the impact of finetuning with LoRA, the effect of encoder size, and the number of ViT-g/14 encoders on binary contact prediction using the Damon dataset. These experiments confirm that LoRA improves all metrics, larger encoders enhance recall and semantic accuracy, and dual encoders offer marginal gains over a shared encoder setup. Based on these findings, our final model uses two ViT-g/14 encoders finetuned with LoRA.

\section{Discussion}

For binary contact prediction, DecoDINO improves the DECO baseline on the DAMON benchmark from an F1 of $56.4 \%$ to $62.5\%$ $(+6.1 \%)$ while cutting the median geodesic error from $18.7$ cm to $15.9$ cm $(–15 \%)$. Ablation showed that replacing DECO's encoders with two ViT-g/14 backbones injected richer global and local cues, lifting both precision and recall. Patch-level attention further reduces geodesic error, confirming that fine-grained token interactions matter for precise contact geometry. Adding a positive class weight $\varphi$ suppresses habitual false-positive foot contacts, trading a small recall drop for large precision and localization improvements. The recall drop after reweighting mirrors our qualitative finding that DECO over-relied on a "feet-on-ground" prior; the new weighting corrects this bias a bit, but occasionally misses rare contact vertices and still fails when there is no person in the scene. Qualitative results also revealed the model still makes various errors with more detailed contact (e.g. fingers and heels). Where the positive class weight allowed us to overcome some bias, DecoDINO still predicts feet contact when there is not even a person in the scene. Qualitatively, we see that the model predicts the contact object reasonably good. The semantic head achieves high recall, meaning it identifies most of the relevant semantic labels, whereas, its low precision indicates many incorrect positive predictions. DecoDINO’s performance indicates that, despite improved performance, the model remains prone to certain failure cases and requires further refinement to improve reliability.

 Including the lightweight SmolVLM enlarges the model but yields no measurable gains, suggesting that visual features already encode sufficient context and that VLM tokens are not well aligned with dense contact geometry.

\subsection{Conclusion}
We investigated DECO's systematic errors and introduced DecoDINO, a contact-aware transformer that couples two LoRA-tuned DINOv2-G encoders with a patch-level cross-attention fusion and a positive class balance loss. On DAMON it raises binary-contact F1 by $6\%$, reduces geodesic error by $2.8$ cm, and delivers the first per-vertex semantic labels in this setting. Ablations confirm that richer ViT features and explicit class re-weighting are the primary drivers of the gain, whereas a compact VLM branch provides no added value. The overall performance indicate that, despite improved performance, the model remains prone to certain failure cases and requires further refinement to improve reliability.

\subsection{Future Work}

Several directions can extend this work. First, increasing semantic recall should be prioritized by adding datasets with denser vertex labels. Second, although our attempt to integrate VLM did not result in better performance, it could still be of great interest. Stronger modalities may be obtained by replacing SmolVLM with a frozen large-scale model (e.g. SigLIP \citep{slip}), while training only lightweight adapters to ensure gradients propagate between vision and language streams. Third, the project's scope forced us to strictly staying within DECO’s framework, which may not be the best option. Ablation showed that
using two ViT-g/14 encoders instead of one yields almost no additional benefit ($\Delta$ F1 = $–0.1 \%$) but doubles inference FLOPs. Instead, we could distill the dual-ViT architecture into a single, medium-sized backbone or a sparse mixture-of-experts. Finally, broader evaluation on benchmarks (e.g. RICH \citep{huang2022capturing} or PROX \citep{hassan2019resolving}), combined with studies of zero-shot transfer to unseen domains via continual self-supervision, will clarify how well DecoDINO generalizes beyond DAMON.

\subsection{Challenges}
During the project, we were tasked with building on DECO, which restricted us from making major architectural changes. This constraint led us to use two separate encoders instead of simplifying the model with a shared one, which could have reduced complexity and possibly improved performance. Furthermore, since the project was part of the RHOBIN challenge, we were limited to using only the DAMON dataset, preventing us from exploring richer alternatives such as PICO-db \citep{cseke2025picoreconstructing3dpeople}.

Another significant limitation was the availability of the Snellius GPU cluster. The cluster experienced several multi-day outages due to maintenance and issue resolution, which restricted access to computational resources. To continue development during downtime, the team set up the codebase for GPU usage on a local machine, allowing some experiments to proceed despite the delays.

In addition, integration of DECO's pixel anchoring renderer on Windows. This required building PyTorch3D from source, which failed due to missing EGL/OSMesa libraries. Resolving these issues took several days of development for team members using Windows. Ultimately, the most efficient solution was to switch to the Linux-based Snellius cluster. This also addressed another constraint, as the datasets were too large to store and process effectively on local machines.

Finally, adding a VLM branch introduced significant complexity. This included caching hidden states, synchronizing text and visual features, and managing LoRA adapters. However, it yielded no performance gain on binary or semantic metrics (Table \ref{tab:vlm}). To understand whether the VLM features were noisy or simply misaligned, it took additional experiments beyond our original timeline.

\subsection{Task Division}
The project tasks were loosely divided among team members across coding, research, qualitative analysis, writing, and poster design. Davide and Lukas were the main people  responsible for implementing the codebase, running experiments, and designing and testing various architectures. They also contributed to the final writing. Fleur was the main person of writing the paper and poster design, including all visualizations. She conducted the qualitative analysis and developed two supporting notebooks. Additionally, she implemented and tested the model on the RICH and PROX datasets, though those results were not included in the final paper. Angelo has set up the qualitative analysis. Helia contributed to the paper writing, poster design, and supported the qualitative analysis. All team members independently conducted research to support the project and its documentation.

{
    \small
    \bibliographystyle{ieeenat_fullname}
    \bibliography{reference}
}
\onecolumn

\noindent\textbf{\large{Appendix}}
\appendix
\section{Qualitative Failure Analysis of DECO}
\label{Qualitative}

\begin{figure}[!h]
\centering
    \includegraphics[width=0.80\textwidth]{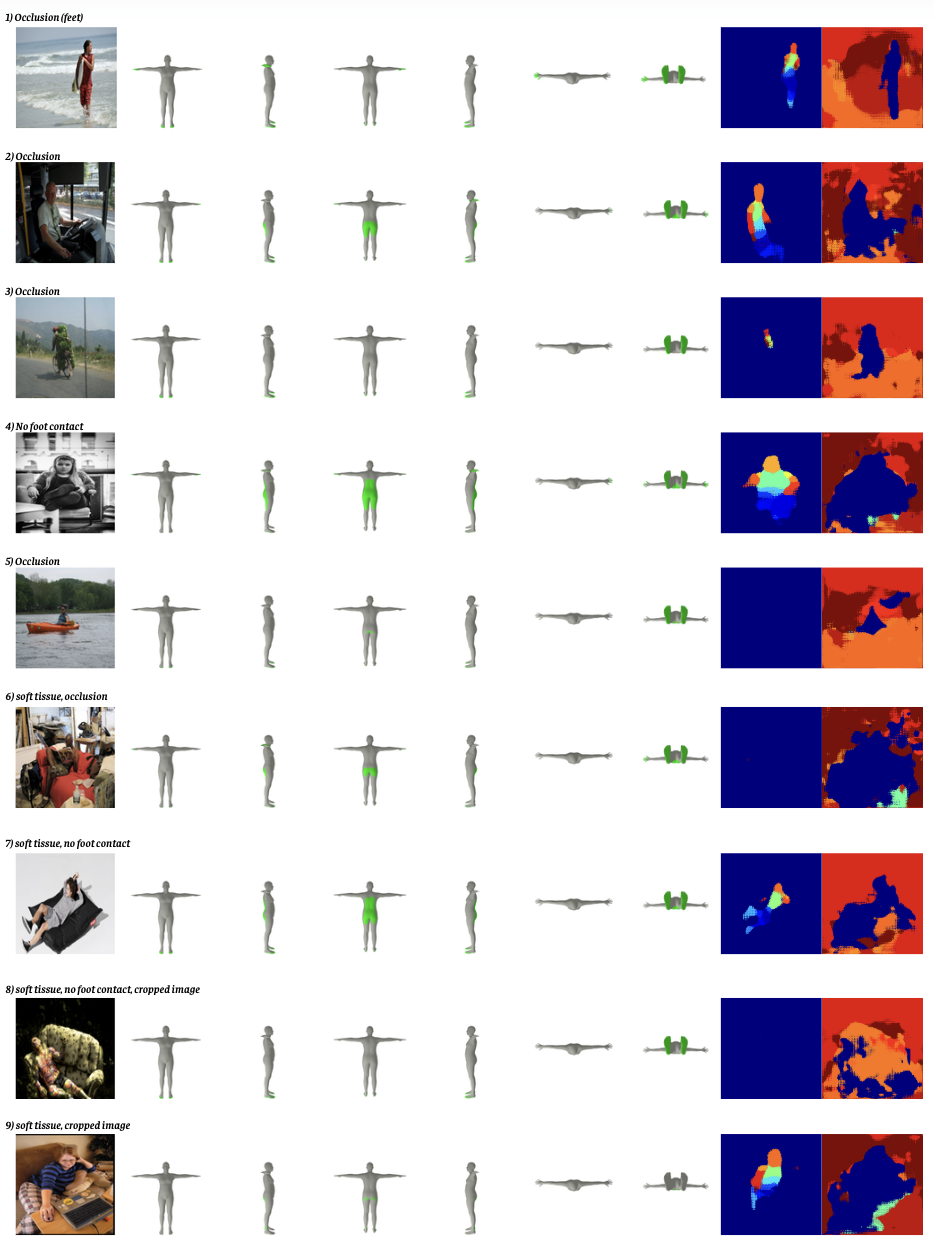}
    \captionof{figure}{Qualitative analysis on challenging tasks (1/2): \textbf{Left.} Original image. \textbf{Middle.} Binary contact prediction on SMPL body mesh from different angles. \textbf{Right.} Part and scene mask of the image, respectively.}
    \label{fig:part1}
\end{figure}

\begin{figure}[!h]
    \centering
    \includegraphics[width=0.8\linewidth]{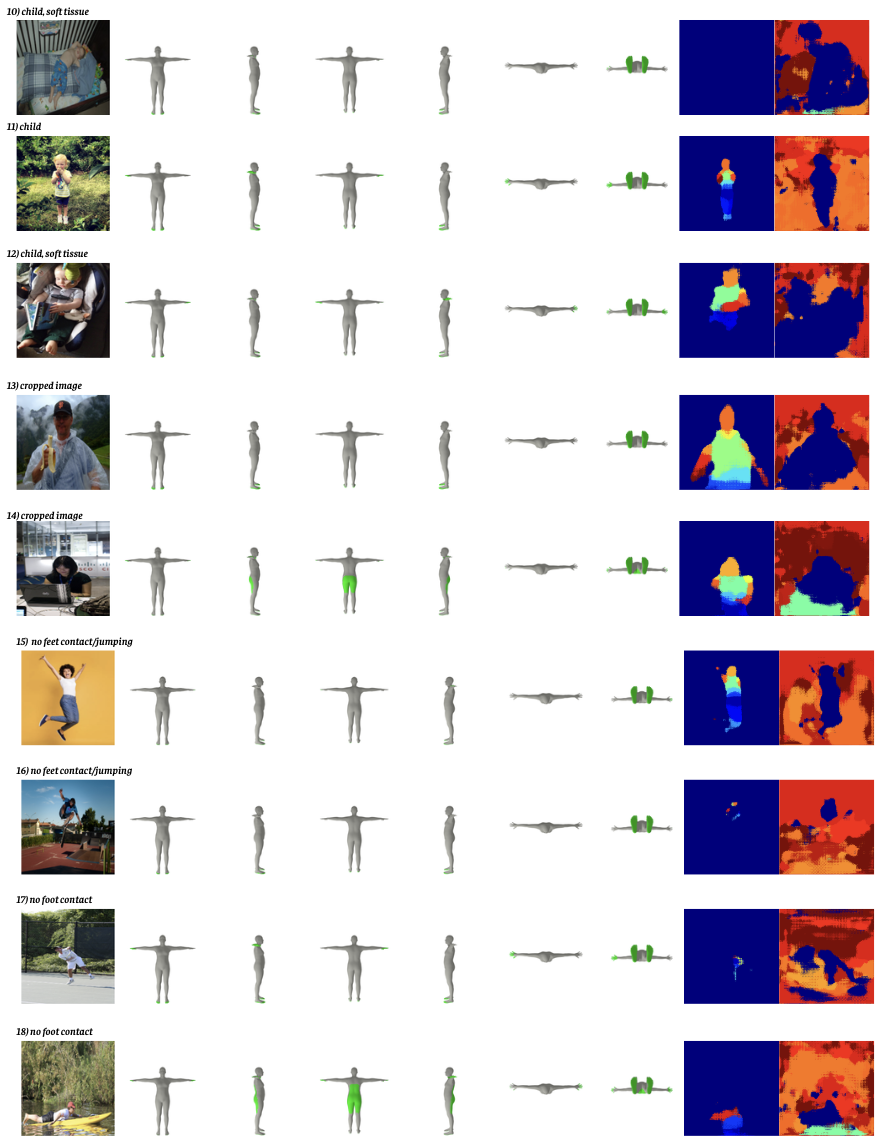}
    \caption{Qualitative analysis on challenging tasks (2/2): \textbf{Left.} Original image. \textbf{Middle.} Binary contact prediction on SMPL body mesh from  different angles. \textbf{Right.} Part and scene mask of the image, respectively.}
    \label{fig:part2}
\end{figure}
\newpage
\twocolumn
\section{DECO }
\label{DECOappendix}

\paragraph{Cross-attention} utilizes the queries, keys and values for the scene-context branch $\{Q_s,K_s,V_s\}= \{F_s,F_s,F_s\}$ and the part-context branch $\{Q_p,K_p,V_p\}= \{F_p,F_p,F_p\}$. It allows us to exchange the $Q$ in the multi-head attention block between the two branches, obtaining the contact features $F_c$
\[
    F'_s = \text{softmax}(\frac{Q_pK_s^T}{\sqrt{C_t}})V_s
\]
\[
    F'_p = \text{softmax}(\frac{Q_sK_p^T}{\sqrt{C_t}})V_p
\]
\[
    F_c = LN(F'_s \odot F'_p)
\]

where $C_t$ is a scaling factor, $\odot $ the Hadamard operator and $LN$ a layer-normalization. $F_c$ is filtered by a shallow MLP followed by sigmoid activation, outputting $\bar{y}_c \in \mathbb{R}^{6890\times 1}$

\paragraph{Loss}
 A $\mathcal{L}_c^{3D}$ is the binary-cross entropy loss between per-vertex predicted contact $\bar{y}_c$ and ground-truth contact labels $y^{gt}_c$:
 \begin{equation}
    \mathcal{L}_c^{3D} = -\frac{1}{N} \sum^N_{i=1} [ \underbrace{y_i\log (p_i)}_{\text{positive term}} + \underbrace{(1-y_i)\log (1-p_i)}_{\text{negative term}}]
    \label{BCE}
\end{equation}

 Additionally, the 2D pixel anchoring loss $\mathcal{L}_{pal}^{2D}$ is used to relate contact on the 3D mesh with image pixels. PAL grounds 3D predictions by (1) estimating camera and SMPL parameters with CLIFF, (2) rendering the colored mesh via a differentiable renderer (PyTorch3D) under weak perspective, and (3) comparing the resultant 2D contact map against crowd-sourced 2D annotations using a binary cross-entropy loss. DECO is trained end-to-end by summing these two losses with two segmentation losses $\mathcal{L}_s^{2D}$ and $\mathcal{L}_p^{2D}$ between the predicted and the ground-truth masks:

\begin{equation}
\mathcal{L} =  w_c\mathcal{L}_c^{3D}+ w_{pal}\mathcal{L}_{pal}^{2D}
+ w_s\mathcal{L}_s^{2D}
+ w_p\mathcal{L}_p^{2D}
    \label{total_deco_loss}
\end{equation}

\section{DINOv2}
\label{DINO}

The image-level DINO loss is:

\[\mathcal{L}_{DINO} = -\sum p_t \log p_s \]

Summing over each masked patch $i$, the iBOT loss term is defined as:

\[\mathcal{L}_{iBOT} = -\sum_i p_{ti} \log p_{si} \]

\noindent Both the $\mathcal{L}_{DINO}$ and $\mathcal{L}_{iBOT}$ train the student network parameters, whereas the teacher parameters are updated through an exponential moving average of the student's parameters, maintaining stability and consistency in the learned representations.

\section{Attention Pooling}
\label{attpooling}
Let $F \in \mathbb{R}^{N \times D}$ denote the feature map output by the cross-attention module, where $N$ is the number of patches and $D$ is the feature dimension. To aggregate these patch-level features into a single feature vector, we use an attention-based pooling mechanism. Specifically, we introduce a learnable query vector $q \in \mathbb{R}^D$ and compute attention scores over all $N$ patches:
\begin{equation*}
    \alpha_i = \frac{\exp(q^TF_i)}{\sum_{j=1}^N(\exp(q^TF_j)}, \quad i = 1,\dots ,N
\end{equation*}
The output is a weighted sum of the patch embeddings:
\begin{equation*}
    F_{att} = \sum_{i=1}^N\alpha_iF_i \in \mathbb{R}^{1xD}
\end{equation*}

\section{VLM}
\subsection{Prompts}
\label{VLM prompts}

 \begin{figure}[!h]
    \centering
    \begin{subfigure}[t]{0.4\linewidth}
        \centering
        \includegraphics[width=3cm]{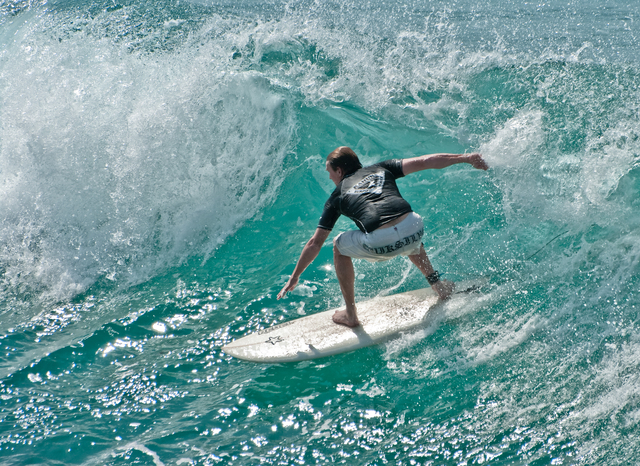}
        \caption{Text prompt: \textit{"The human is in contact with a surfboard, which is being used to ride a wave. The human is performing the action of surfing."}}
        \label{input}
    \end{subfigure}%
    \quad
    \begin{subfigure}[t]{0.4\linewidth}
        \centering
        \includegraphics[width=3cm]{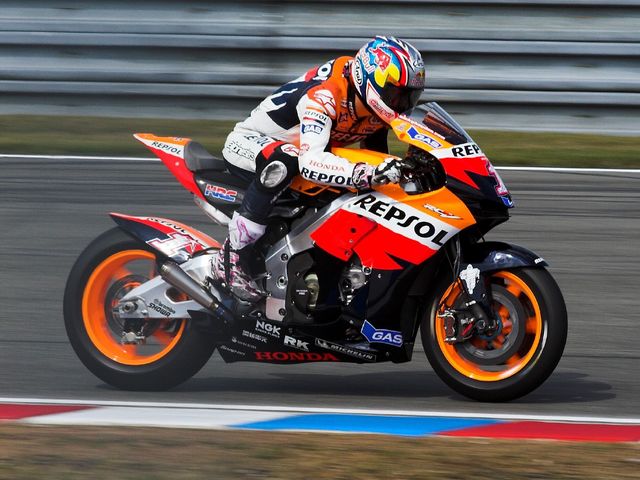}
        \caption{Text prompt: \textit{"The human is in contact with the motorcycle, which is in motion. The human is wearing a helmet, gloves, and a jacket. The human is in contact with the motorcycle's handlebars, which are being held with the left and right hand. The human is in contact with the motorcycle's seat, which"}}
        \label{k0}
    \end{subfigure}%

    \caption{Two example images with corresponding VLM prompts.}
     \label{fig:prompts}
\end{figure}

\subsection{During inference}
\label{VLM inference}

\begin{enumerate}
    \item We tokenize the prompt and attach the single RGB image as a "visual input" to SmolVLM. 
    \item We run \texttt{generate(...)} with a maximum of 60 new tokens. As the VLM generates each token, out hook collects the corresponding hidden vector. 
    \item At the end of generation, we receive token-level outputs describing the scene in the image. We pass these through the text bidirectional text encoder to process them, resulting in a dense sequence of vectors $T_{\text{img}}\in\mathbb{R}^{N_t\times D_h}$, where $N_t$ is the number of generated tokens and $D_h$ is the encoder hidden dimension.
    \item For efficiency, we store $T_{\text{img}}$ on disk, hashed by image filename, so that repeated passes over the same image reuse cached features rather than recompute VLM outputs.  
\end{enumerate}

\section{Ablation Studies}
\label{ablation}

\paragraph{Finetuning strategy.} We test whether lightweight finetuning with LoRA boosts binary contact prediction performance, using a single ViT-g/14 encoder for both scene and part branches.  Enabling LoRA improves every metric, most notably lowering the  geodesic error with $- 3.02$cm. 


\begin{table}[!h]
\resizebox{\linewidth}{!}{%
    \begin{tabular}{ccccc}
    \toprule
    &\multicolumn{4}{c}{\textbf{Binary Contact Prediction}}\\
      \textbf{Finetuning} & F1$_{\%}$ & Precision$_{\%}$ & Recall$_{\%}$ & Geo. error$_{cm}$ \\
     \midrule
    $ \times$ &  62.69   &  56.70   &   81.32  &  25.83 \\
    \checkmark & \textbf{63.99}   &  \textbf{58.55}   &   \textbf{81.54}  &  \textbf{22.81} \\
    \bottomrule
    \end{tabular}
    }
    
    \caption{Ablating the use of LoRA for finetuning.}
\end{table}

\paragraph{Encoder size.} To see the benefit of a larger image backbone, we swap ViT-L/14 for ViT-g/14 while keeping all hyper-parameters fixed and both finetuned with LoRA. ViT-g adds 66 \% more parameters (1.1 B vs.\ 0.67 B) but yields the best F1 (+0.97 \%) and recall (+3.3 \%) for binary contact prediction; ViT-L achieves the lowest geodesic error.

\begin{table}[!h]
\resizebox{\linewidth}{!}{%
    \begin{tabular}{lcccc}
    \toprule
    &\multicolumn{4}{c}{\textbf{Binary Contact Prediction}}\\
      \textbf{Type}& F1$_{\%}$ & Precision$_{\%}$ & Recall$_{\%}$ & Geo. error$_{cm}$ \\
     \midrule
    ViT-L/14& 62.94  &   \textbf{58.86}  &  78.31   &  \textbf{20.91} \\
    
     ViT-G/14&  \textbf{63.91}   &  58.44   &   \textbf{81.63}  &  22.17 \\
     \bottomrule
    \end{tabular}
    }
    \caption{Ablating the encoder size: Large vs Giant.}
\end{table}

     

\paragraph{Number of encoders.} We ablate the number encoder between using either one or two ViT-g/14 encoder, while both using LoRA for finetuning. For the single-encoder run we feed the same features to both scene and part branches, which have shared weights, whereas the double-encoder has separate weights. Both variants have separate cross-attention as in original DECO. 

\begin{table}[!h]
\resizebox{\linewidth}{!}{%
    \begin{tabular}{ccccc}
    \toprule
    &\multicolumn{4}{c}{\textbf{Binary Contact Prediction}}\\
     \textbf{\# Enc.} & F1$_{\%}$ & Precision$_{\%}$ & Recall$_{\%}$ & Geo. error$_{cm}$\\
     \midrule
    1& \textbf{63.99}   &  \textbf{58.55}   &   81.54  &  22.81 \\
     
     2&  63.91   &  58.44   &   \textbf{81.63}  &  \textbf{22.17} \\
     \bottomrule
    \end{tabular}
    }
    \caption{Ablating the number of encoders.}
\end{table}

     
Using two independent encoders shows slightly better binary contact prediction performances on recall, geodesic error and semantic accuracy, however, these performance differences are minimal.

\section{Qualitative Results}
\label{app:qualitative results}
\begin{figure*}[!h]
    \centering
    \includegraphics[width=0.9\linewidth]{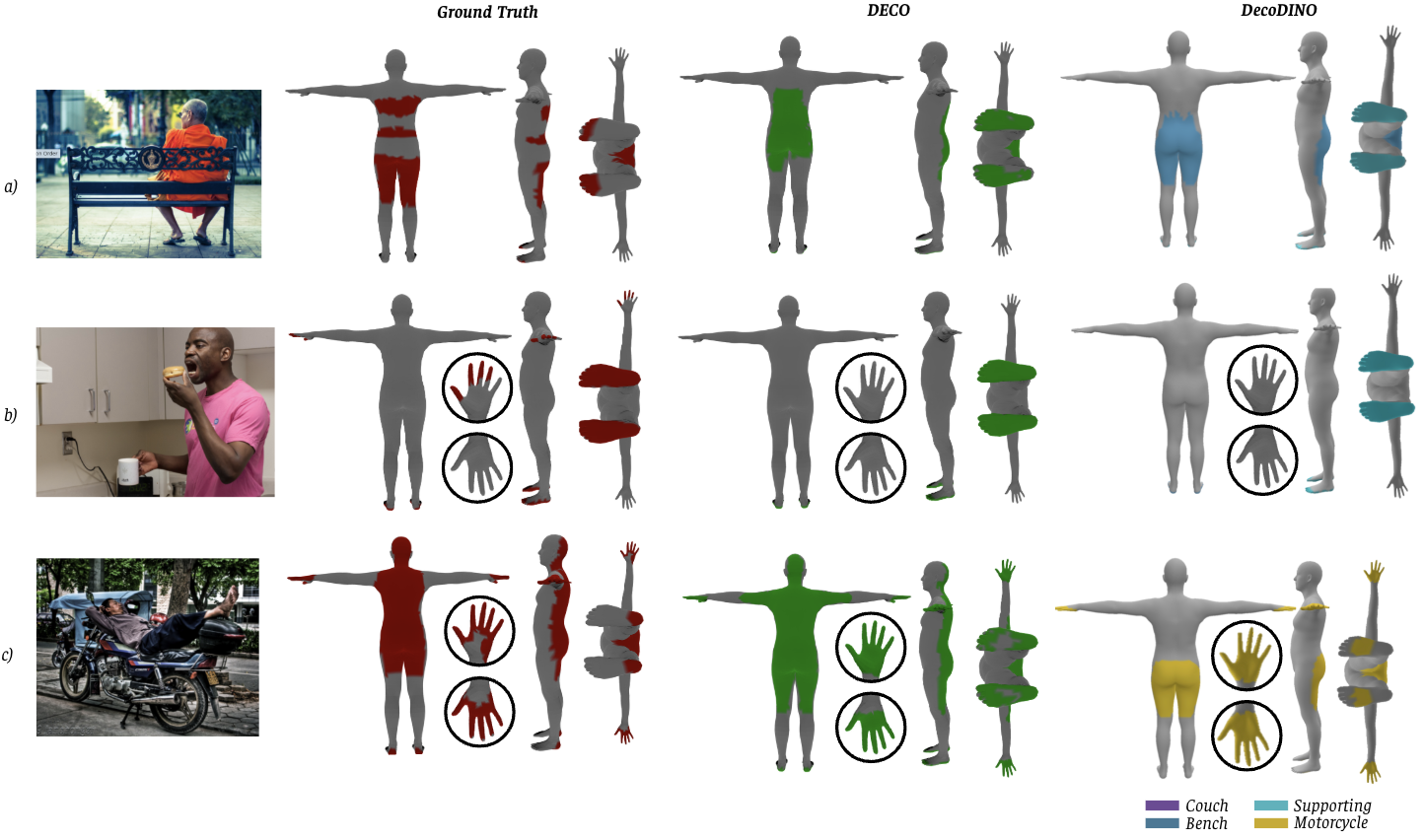}
    \caption{\textbf{Qualitative Results}. (a) DECO underestimates contact, particularly on the legs and back, but over estimates on the back. DecoDINO captures a more complete contact pattern on the legs but. misses some contact regions of the back. DecoDINO correctly labels the object as a bench (blue). (b) DECO and DecoDINO both correctly predict foot contact, but miss contact prediction on the finger. However, both models over-predict contact on the feet and under-predict contact on the hands, where DecoDINO assigns 'supporting' as semantic label (light blue). (c) DECO overestimates on the upper-arms and feet, whereas DecoDINO, localizes the contact primarily to the lower body but misses te whole back, head and also predicts incorrect regions of the feet. It does assigns the correct object label motorcycle (yellow).}
    \label{fig:results_extra}
\end{figure*}


\end{document}